\documentclass[conference]{IEEEtran}
\IEEEoverridecommandlockouts
\usepackage{cite}
\usepackage{amsmath,amssymb,amsfonts}
\usepackage{algorithmic}
\usepackage{graphicx}
\usepackage{textcomp}
\usepackage{xcolor}
\usepackage{booktabs} % 提供 \toprule, \midrule 等命令
\usepackage{multirow} % 提供 \multirow 命令

\def\BibTeX{{\rm B\kern-.05em{\sc i\kern-.025em b}\kern-.08em
    T\kern-.1667em\lower.7ex\hbox{E}\kern-.125emX}}

\newcommand{\refsec}[1]{Section~\ref{sec:#1}}

\begin{document}

\title{EasyGenNet: An Efficient Framework for Audio-Driven Gesture Video Generation Based on Diffusion Model}

% \author{Anonymous ICME submission}
\author{
    \IEEEauthorblockN{
Renda Li\textsuperscript{1},
Xiaohua Qi\textsuperscript{1},
Qiang Ling\textsuperscript{1},
Jun Yu\textsuperscript{1}\textsuperscript{*},
Ziyi Chen\textsuperscript{2},
Peng Chang\textsuperscript{2},
Mei Han\textsuperscript{2}
Jing Xiao\textsuperscript{2}
}

\IEEEauthorblockA{\textsuperscript{1}University of Science and Technology of China, Hefei, China}

\IEEEauthorblockA{\textsuperscript{2}PAII Inc.}
}

\maketitle

% \begin{figure*}[t]
%   \centering
%   \includegraphics[width=\linewidth]{images/teaser.pdf}
%   \caption{Given an audio segment, a speaker ID, and a reference image of the speaker, our co-speech gesture video generation framework can produce videos with realistic appearances, vivid facial expressions, and clear hand gestures. Here, we showcase several generated video frames for two characters.}
%   \label{fig:teaser}
% \end{figure*}

\begin{abstract}
 Audio-driven cospeech video generation typically involves two stages: speech-to-gesture and gesture-to-video. While significant advances have been made in speech-to-gesture generation, synthesizing natural expressions and gestures remains challenging in gesture-to-video systems. 
In order to improve the generation effect, previous works adopted complex input and training strategies and required a large amount of data sets for pre-training, which brought inconvenience to practical applications.
We propose a simple one-stage training method and a temporal inference method based on a diffusion model to synthesize realistic and continuous gesture videos without the need for additional training of temporal modules.
The entire model makes use of existing pre-trained weights, and only a few thousand frames of data are needed for each character at a time to complete fine-tuning.
Built upon the video generator, we introduce a new audio-to-video pipeline to synthesize co-speech videos, using 2D human skeleton as the intermediate motion representation. 
Our experiments show that our method outperforms existing GAN-based and diffusion-based methods. 
\end{abstract}


\section{Introduction}
\label{sec:intro}

Co-speech video generation~\cite{ginosar2019gestures} focuses on creating photorealistic videos of talking humans that are closely synchronized with the input speech. Such videos not only match the lip movements to the spoken words but also incorporate accurate facial expressions and gestures.

Co-speech video generation usually involves two stages: speech-to-gesture and then gesture-to-video. Significant advances have been made in speech-to-gesture generation, exemplified by the work in ~\cite{yi2023generating} where a holistic 3D mesh-based human motion sequence can be generated from a given speech. However, most gesture-to-video works rely on large-scale pre-training \cite{makeyouranchor, vlogger, cyberhost}, and multi-stage training \cite{animateanyone, magicanimate, magicdance2023}, which makes it difficult to quickly apply to characters outside the training set. We hope to design a model that can quickly finetune on a specific person while maintaining the generated visual quality. GAN-based models have advantages in training and inference speed, but previous methods \cite{ginosar2019gestures} cannot generate realistic gesture videos. Therefore, we use the latest GAN-based methods \cite{myeccv2024} to build a baseline and thoroughly analyze the disadvantages of GAN in generalization ability compared with diffusion models. GAN-based methods do not extrapolate well to unseen gestures and especially have difficulty in generating novel hand poses, resulting in blurry or missing fingers in certain cases. Finally, we design a diffusion model-based framework that fully utilizes the existing pre-trained weights, so that only a single-stage finetune and a few thousand frames of character videos are needed to quickly fine-tune the video generation model for a specific person.

\begin{figure}[t!]
  \centering
  \includegraphics[width=\linewidth]{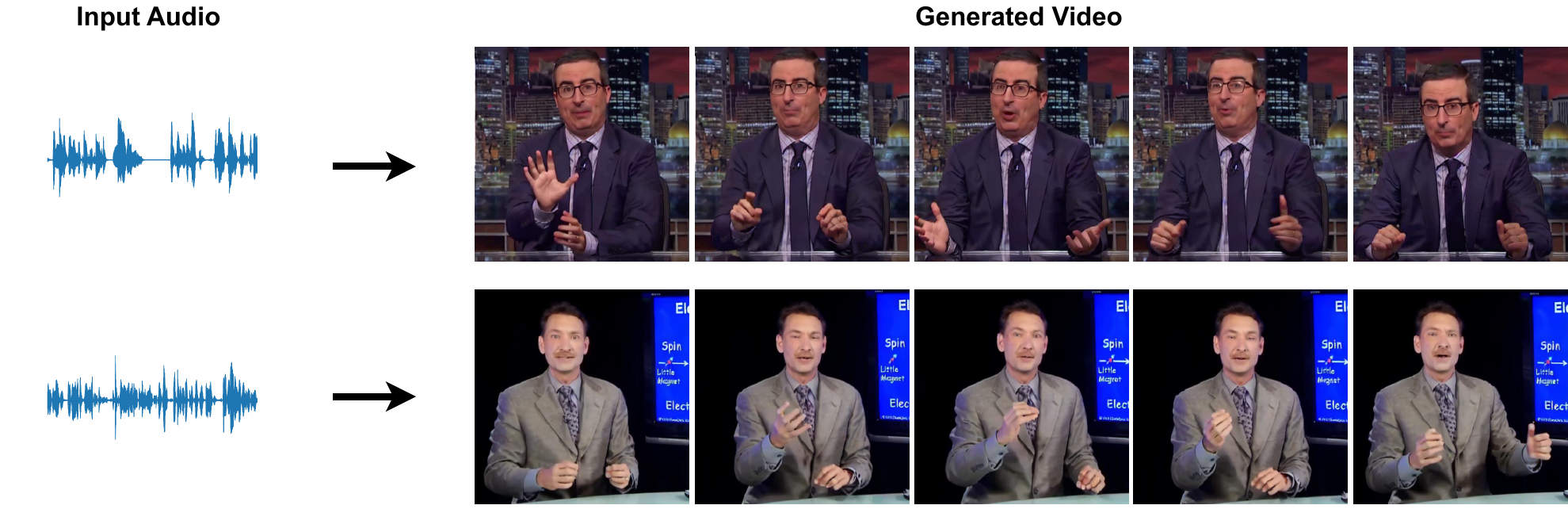}
  \vspace{-10pt}
  \caption{Given an audio segment, a speaker ID, and a reference image of the speaker, our co-speech gesture video generation framework can produce videos with realistic appearances, vivid facial expressions, and clear hand gestures. Here, we showcase several generated video frames for two characters.}
  \label{fig:teaser}
\vspace{-10pt}
\end{figure}

% \begin{teaserfigure}
%     \centering	
%     \includegraphics[width=\linewidth]{images/teaser.pdf}
%     % \vspace{-0.7cm}
%     \caption{
%     xxxxx
%     }
%     \label{fig:teaser}
% \end{teaserfigure}

In this work, we present an audio-to-video pipeline to generate co-speech videos of selected speakers, with realistic facial animation and novel hand gesture movements. To minimize the training difficulty, we introduce a simple SD-based Network, referred to as EasyGenNet, to enable the synthesis of human hand movements in gesture videos with significantly enhanced visual quality. Specifically, we propose to generate the 2D skeleton maps from a 3D mesh-based human motion sequence, explicitly enforcing the 3D human body physical constraint and resulting in improved synthesis of the human body, especially for hands. At the same time it allows to train on small datasets with a pre-trained ControlNet~\cite{controlnet}. Finally, we implement temporal inference by performing self-attention on the temporal dimension, so that continuous and smooth videos can be generated even after removing the temporal module.

Our main contributions include the following:
\begin{itemize}

    \item We propose a simple SD-based co-speech gesture video generation pipeline (EasyGenNet), trained on a modest amount of data and without the need for complex input and multi-stage training strategies, to generate satisfactory human speech video with realistic gestures and hand motions.
    \item We use the 3D human body parametric model SMPLX to obtain more accurate 2D skeleton images, significantly improving the SD synthesis result, most notably for human hands. Our method can be easily combined with the existing Audio-to-SMPLX method to expand its application capabilities. 
    \item We discuss the advantages of diffusion-based models over the state-of-art GAN-based models through quantitative and qualitative experiments, and and achieves better performance than other diffusion-based methods without requiring complex model and training strategies.

\end{itemize}

% \begin{figure}[t!]
%   \centering
%   \includegraphics[width=\linewidth]{images/teaser.pdf}
%   \vspace{-10pt}
%   \caption{Given an audio segment, a speaker ID, and a reference image of the speaker, our co-speech gesture video generation framework can produce videos with realistic appearances, vivid facial expressions, and clear hand gestures. Here, we showcase several generated video frames for two characters.}
%   \label{fig:teaser}
% \vspace{-10pt}
% \end{figure}

\section{Related work}
\label{sec:related_works}

\subsection{Co-speech Human Motion Generation.}
Research in co-speech motion generation has largely been divided into two categories: 
facial motion~\cite{cudeiro2019capture, karras2017audio, richard2021meshtalk,fan2022faceformer} and body motion~\cite{yi2022generating, zhu2023taming, Ao2023GestureDiffuCLIP, liu2022audio}. Despite these advancements and recent efforts, creating a unified framework that seamlessly integrates the generation of facial expressions, body movements, and hand gestures remains difficult. 
Recently, several works have tackled the above challenge by leveraging human body parametric models~\cite{li2017learning, boukhayma20193d, loper2023smpl, pavlakos2019expressive}. 
More recently, Yi et al.~\cite{yi2022generating} propose to generate full-body 3D meshes using an encoder-decoder model for facial motions and a cross-model VQ-VAE for modeling body and hand motion. Our study extends this methodology and takes a step further to generate co-speech videos.

\subsection{Gesture Video Generation.}
GAN-based methods \cite{ginosar2019gestures, chan2019dance,wang2018vid2vid} learn to predict 2D skeleton-based gestures given an input speech and use conditional GANs to generate final videos. The recent state-of-art GAN-based method \cite{myeccv2024} uses a textured mesh as a stronger condition. We use the framework of \cite{myeccv2024} as a baseline and find that GAN-based video generators often fail to generate clear hands and generalize to unseen poses. Some recent work based on diffusion models \cite{magicanimate, magicdance2023, animateanyone, champ, makeyouranchor, vlogger} has alleviated the problems of GAN-based models to some extent. However, \cite{makeyouranchor} requires 27 hours of pre-training to adapt the model to 2D mesh representation, and requires additional face enhancement to improve facial clarity. \cite{vlogger} focuses on zero-shot capabilities, so it uses 2.2k hours of data pre-training. 
The 2D skeleton images in \cite{animateanyone, magicdance2023} are directly extracted using models like OpenPose, which may lead to missing hand, while \cite{champ} relies on non-direct extraction from 3D meshes, potentially causing inconsistencies. 
Our method generates realistic character videos in a fine-tuned manner without the need for additional pre-training and temporal module training. The 2D skeleton map used is directly projected from 3D keypoints, avoiding missing and inaccurate hands.

\section{Method}
\label{method}

\begin{figure}[t!]
\centering %
\includegraphics[width=0.8\linewidth]{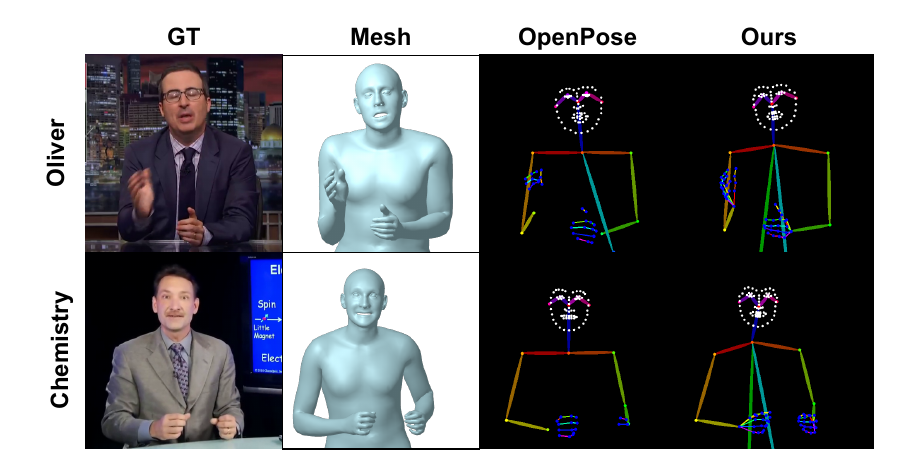} 

\vspace{-10pt}
\caption{
 Above chart illustrates skeleton maps generated with two methods: inferring GT images with OpenPose model ($3$rd column), and our method which extracts SMPLX model parameters then mapping them into joints under the OpenPose configuration ($4$th column). Our method yields more accurate hand poses, finger configurations and body shapes.
}
\label{fig:3_mesh_skeleton}
\vspace{-10pt}
\end{figure}

Our goal is to generate an audio-driven, photorealistic, co-speech gesture video from a segment of audio given a speaker ID and a reference image of a speaker. We address this challenge in two steps. First, given an audio, we generate a sequence of plausible body poses and facial expressions. Next, we use the generated sequence of 2D skeleton poses and a reference image to condition an SD-based pipeline to generate the photorealistic video. We have described the first and second steps in more detail in \ref{sec:audio2gesture} and \ref{sec:gesture2video}, respectively.
% \input{figures/2_method}
% \begin{figure*}[t!]
%     \centering
%     \includegraphics[width=0.99\linewidth]{images/Method_3.pdf}
%     % \vspace{-2mm}
%     \caption{
%     Given an input audio $\mathcal{A}_t$, we first train a network to predict skeleton sequences that match speech, as denoted as $\mathcal{P}_t$ for each frame in \refsec{audio2gesture}. Additionally, leveraging the pre-trained SD 1.5 model and Pose ControlNet, we devise a multi-stage fine-tuning video generation model. Given a skeleton sequence $\mathcal{P}_t$ and a reference image as conditions, our model generates videos that align with the appearance of the reference image while matching the poses to the skeleton sequence.
%     }
%     \label{fig:method}
% % \vspace{-2mm}
% \end{figure*}

\begin{figure*}[t!]
    \centering
    \includegraphics[width=0.9\linewidth]{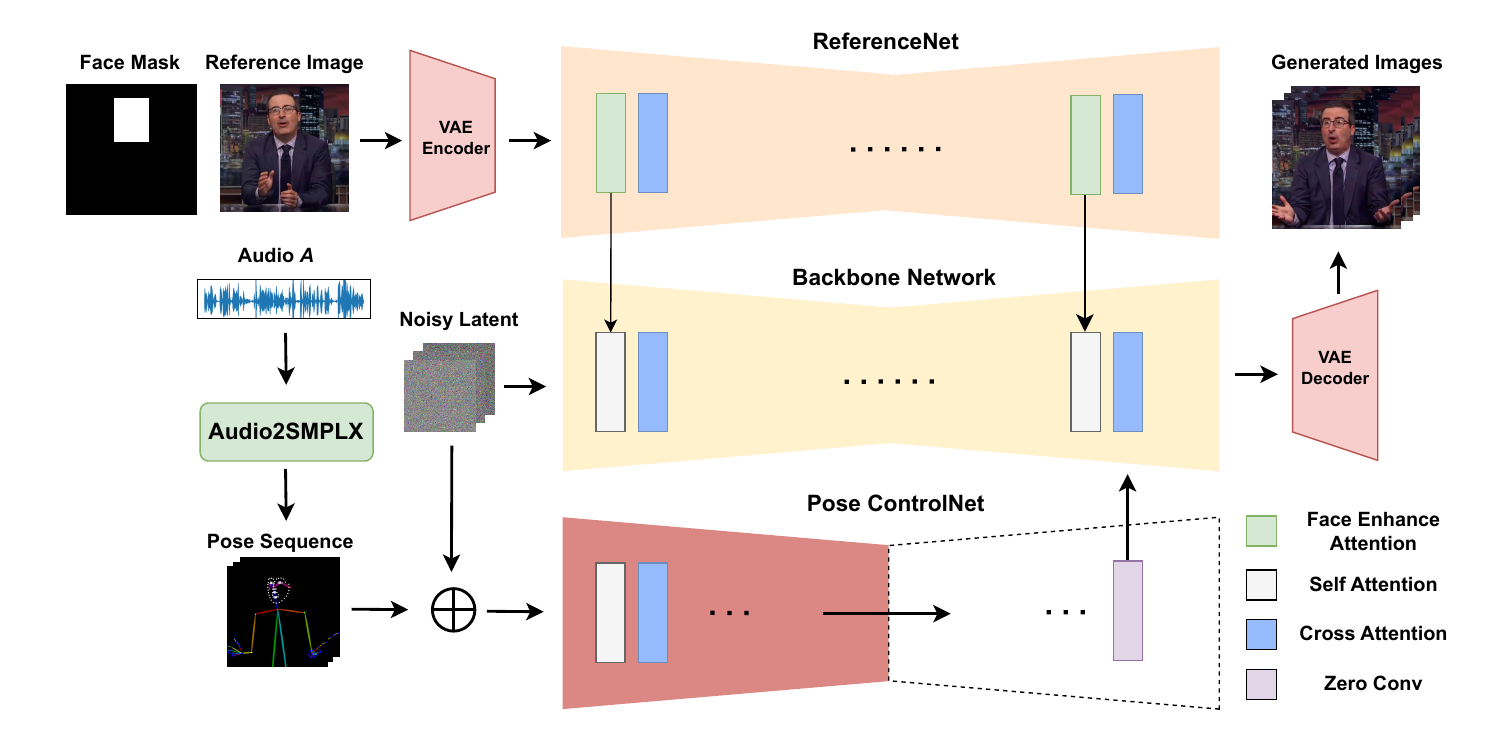}
    
        \vspace{-10pt}
    \caption{
    Given an input audio $\mathcal{A}_t$, we first train a network to predict skeleton sequences that match speech, as denoted as $\mathcal{P}_t$ for each frame in \refsec{audio2gesture}. Additionally, leveraging the pre-trained SD 1.5 model and Pose ControlNet, we designed a single-stage fine-tuned video generation model called EasyGenNet. Given a skeleton sequence $\mathcal{P}_t$ and a reference image as conditions, our model generates videos that align with the appearance of the reference image while matching the poses to the skeleton sequence.
    }
        \vspace{-10pt}
    \label{fig:method}
\end{figure*}

% \begin{figure}[t!]
%     \centering
%     \includegraphics[width=0.7\linewidth]{images/Method_7.pdf}
%         % \vspace{-10pt}
%     \caption{
%     Given an input audio $\mathcal{A}_t$, we first train a network to predict skeleton sequences that match speech, as denoted as $\mathcal{P}_t$ for each frame in \refsec{audio2gesture}. Additionally, leveraging the pre-trained SD 1.5 model and Pose ControlNet, we designed a multi-stage fine-tuned video generation model called CoGenNet. Given a skeleton sequence $\mathcal{P}_t$ and a reference image as conditions, our model generates videos that align with the appearance of the reference image while matching the poses to the skeleton sequence.
%     }
%         % \vspace{-10pt}
%     \label{fig:method}
% \end{figure}

\subsection{Audio to Gesture}
\label{sec:audio2gesture}
Given an audio input, we first predict plausible motion sequence of hands, face and body. We use human skeleton maps in OpenPose configuration as our gesture representation. We use TALKSHOW~\cite{yi2023generating} to generate SMPLX~\cite{pavlakos2019expressive} parameters from audio and then extract the 3D joints and project them into a 2D skeleton image based on the camera parameters of the training video. We use the SMPLX as an intermediate step as it prevents generating unnatural gestures and gives more temporally consistent results.

Given an input audio and a speaker ID, the model predicts the positions of each 3D joint of every single frame. Those positions are then mapped to the configuration of 135 joints utilized by OpenPose. Following this, the camera settings from the source video are used to project those joints onto a 2D plane. The projection process generates a sequence of skeleton maps, represented as $\mathcal{P}_{1:T}=\{\mathcal{P}_t\}_{t=1}^{T}$, where $t=\{1...T\}$, and $T$ represents the total number of skeleton maps in the sequence, as illustrated in Figure \ref{fig:3_mesh_skeleton}.

% \myparagraph{From SMPLX Meshes to 2D Skelen Maps}
To generate a sequence of 2D skeleton maps driven by audio, a straightforward approach is to train a model to regress the positions of 2D joints as configured in OpenPose. However, as shown in Figure \ref{fig:3_mesh_skeleton}, OpenPose predictions can have missing and misaligned joints,which are critical for gesture generation. Therefore, we constrain the 2D joint positions using the 3D mesh representation of the SMPLX parametric human model, resulting in more accurate joint positions and body postures.

% we use the estimated 3D shape parameters obtained by fitting the parametric
% body model to the input image

\subsection{Gesture to Video}
\label{sec:gesture2video}
Given the generated 2D skeleton sequence $\mathcal{P}_{1:T}$ and a reference image, our objective is to generate a photorealistic speaker video.

% \myparagraph{Challenges.}
\subsubsection{Challenges}
Our original intention was that the model could be quickly fine-tuned on characters outside the training set. GAN-based models have great advantages in training and inference speed. Therefore, we initially used the model in \cite{myeccv2024} and the proposed refined 2D skeleton map as conditional input. However, due to limited training data, GAN-based methods tend to learn average color for specific body regions. This results in artifacts around body parts with greater motion, such as hands. It is also challenging for GAN-based methods to transfer gestures from one speaker to another. Since GAN-based methods fail to generate hand gestures that have not appeared in the training data.

Most diffusion model-based methods require additional training of a temporal module or a facial enhancement module \cite{makeyouranchor}, as well as large-scale pre-training to adapt the model to a certain representation. To address these challenges, we use the previously mentioned refined 2D skeleton map as condition, which can obtain more continuous results and clearer facial expressions. And we use temporal inference to further enhance the coherence of the generated video. Finally, all our modules do not require specific pre-training or training of additional modules.

% \myparagraph{Overview.}
\subsubsection{Overview}
The overview of our EasyGenNet is depicted in Figure \ref{fig:method}. The \textbf{Backbone Network} takes noisy latent representation as input and predicts the noise introduced at time step T. To preserve the appearance information of the speaker, the \textbf{ReferenceNet} uses the same SD 1.5 structure as the Backbone Network. Given a specific reference image of a speaker, each layer extracts appearance features and input into the corresponding layer of the denoising network. For controlling the poses of the generated individuals, the \textbf{Pose ControlNet} utilizes the original ControlNet\cite{controlnet} structure and loads the ControlNet OpenPose weights. It takes a sequence of 2D skeleton maps as input and extracts features injected into the backbone network.
%

% \myparagraph{Backbone Denoising Network.}
\subsubsection{Backbone Denoising Network}
In the original SD 1.5, each layer's transformer block consists of a self-attention block and a cross-attention block. We are inspired by the setup in \cite{magicdance2023, magicanimate}. In self-attention, concat the output of the corresponding layer of ReferenceNet. The cross-attention operates with tmpty text, so that the entire backbone network can be frozen during training to improve efficiency, unlike \cite{emo} where the cross-attention layer still needs to be trained.

% \myparagraph{ReferenceNet.}
\subsubsection{ReferenceNet}
Effective appearance control is a crucial capability of video generation models, and recent works \cite{emo, animateanyone, magicdance2023, disco2023} have explored various structures for appearance control effects, including ControlNet-like structures, CLIP encoder, and SD-like structures. Experimental results \cite{animateanyone, zhu2023tryondiffusion} demonstrate the importance of employing a structure similar to the Backbone Network for maintaining consistency in character appearance. Our ReferenceNet adopts the structure and pre-trained weights of SD 1.5. Given a feature map $\mathbf{X} \in \mathbb{R}^{b \times h \times w \times c}$ from a layer of ReferenceNet, features $\mathbb{R}^{b \times ( h \times w ) \times c}$ are obtained after passing through a self-attention layer. Different from above methods, our referencenet additionally input a face mask map. In the self-attention, the face tokens are multiplied by a learnable magnification factor greater than 1 to enhance the visual effect of the face without the need for an additional face enhancement module. We name this self-attention Face Enhancement Attention.

% \myparagraph{Pose ControlNet.}
\subsubsection{Pose ControlNet}
ControlNet \cite{controlnet} has been shown to be able to robustly impose multiple types of image control in T2I models. Therefore, we use the pre-trained OpenPose ControNet model, whose output blocks are added to the input latent of the corresponding out blocks of the Backbone Network. The weights of Pose ControlNet are also frozen.

\subsubsection{Training Strategies}
During training, the model takes a single frame image as input, including a 2D skeleton map, its corresponding RGB image, and a randomly selected reference image. Note that, as described in \ref{sec:audio2gesture}, the skeleton maps are obtained by fitting the SMPLX parametric human model, resulting in more accurate joints position. Backbone Network and ReferenceNet are initialized with the same SD 1.5 weight. During training, only the weights of ReferenceNet are fine-tuned.

\subsubsection{Inference Strategies}
Since our method does not use the temperal module to enhance the consistency of the temporal dimension, we use the All-frames Attention in \cite{makeyouranchor} to enhance the temporal modeling during inference. Different from \cite{makeyouranchor}, we do not use overlapping windows during inference, because it cannot significantly improve the continuity between windows. Instead, we use the same noise to initialize different window inputs, which can enhance the continuity and avoid the huge inference time caused by overlapping windows.

\section{Experiments}
\label{sec:experiments}

% \myparagraph{Datasets.}
\subsection{Datasets.}
\label{sec:datasets}
We leverage the speaker dataset provided by TalkSHOW~\cite{yi2022generating}, specifically focusing on the identities of three characters: ``Oliver,'' ``Chemistry,'' and ``Conan. For the audio-to-gesture component, we employ the complete datasets of these three characters for about $27$ hours. In the gesture-to-video phase, given that the dimensions of image frames significantly surpass that of the SMPLX \cite{pavlakos2019expressive} model parameters used in the audio-to-gesture stage, we restrict our training set to approximately $5$K frames from a single video for each character. 
% The frame rate of each video is 30 FPS, with a resolution of 720 $\times$ 1280.
The video frames are resized to a resolution of 512 $\times$ 512. Additionally, we establish a perspective camera for each cropped video of the three distinct speakers, facilitating the projection of 3D skeletons outputted by Sec \ref{sec:audio2gesture} onto the corresponding locations in the cropped images.

We divide the test set into \textbf{in-domain data} and \textbf{out-of-domain data}, each comprising approximately 10\% of the number of frames in the training set, and subject them to the same cropping process as the training set. In-domain data means that it belongs to the same video as the training set. Out-of-domain data represents videos different from training videos. Out-of-domain data exhibits significant body shifts and novel hand poses, posing a stringent test on the model's capacity for generalization.

% \myparagraph{Baseline Model.} 
\subsection{Baseline Model.} 

Here we built a GAN-based model with a similar structure to \cite{myeccv2024}, comprising an 8-layer UNet-like convolutional neural network generator network and a multi-scale PatchGAN architecture discriminator. The total loss consists of the adversarial loss from the discriminator, the color loss, the perceptual loss, and the feature matching loss.

% \myparagraph{Evaluation Settings.}
\subsection{Evaluation Settings.}
Evaluation is conducted through two different approaches. The first approach assesses the performance of our proposed gesture-to-video generation model (EasyGenNet). In this setting, inference is performed using ground-truth skeleton maps for in- and out-of-domain test sets. We employ the SSIM \cite{ssim}, PSNR\cite{psnr}, and LPIPS \cite{lpips} metrics for evaluating the image-level generation quality and FVD \cite{fvd} for evaluating the quality of the generated video. The second approach qualitatively evaluates the generative quality of the entire audio-to-video system. Due to space constraints, the audio-to-video results and video results are provided in the \textbf{supplementary material}. For evaluation, we disregard background effects since the background changes over time in the ground-truth frames. We focus on evaluating the quality of the avatars. 
In the experiments, we utilize RVM ~\cite{lin2022robust} to mask out the background.

\subsection{Gesture to Video}
\label{gesture2video}
We assess video generation quality using ground-truth 2D skeleton maps from both in-domain and out-of-domain test sets. These maps are obtained by estimating SMPL-X model parameters as outlined in Section \ref{sec:audio2gesture} and mapping the joints to match OpenPose configurations. As previously noted, out-of-domain data show greater body shifts and more varied hand poses than training data, highlighting the need for models with robust generalization abilities to produce high-quality videos.

\begin{table}[t!]
    \centering
    \resizebox{0.95\linewidth}{!}{
    
    \begin{tabular}{l c c c c c}
        \toprule

        \textbf{Speaker} & \textbf{Method} & \textbf{SSIM} $(\uparrow)$ & \textbf{PSNR} $ (\uparrow)$ & \textbf{LPIPS} $(\downarrow)$ & \textbf{FVD} $(\downarrow)$
        
        \\
        \cmidrule(lr){1-6}
        \multirow{4}{*}{Oliver}
        & Animate-Anyone & $0.701$ & $29.484$ & $0.302$ & $15.776$ \\
        \cmidrule(lr){2-6}
        & MYA & $0.752$ & $31.871$ & $0.223$ & $13.064$ \\
        \cmidrule(lr){2-6}
        & GAN Baseline & $0.734$ & $31.780$ & $0.258$ & $14.930$ \\
        \cmidrule(lr){2-6}
         & Ours & $\mathbf{0.758}$ & $\mathbf{32.688}$ & $\mathbf{0.219}$ & $\mathbf{12.178}$ \\
         \cmidrule(lr){1-6}
        \multirow{4}{*}{Chemistry}
        & Animate-Anyone & $0.692$ & $30.053$ & $0.329$ & $14.385$ \\
        \cmidrule(lr){2-6}
        & MYA & $\mathbf{0.751}$ & $\mathbf{32.064}$ & $0.269$ & $10.255$ \\
        \cmidrule(lr){2-6}
        & GAN Baseline & $0.741$ & $31.755$ & $0.264$ & $11.237$ \\
        \cmidrule(lr){2-6}
         & Ours & $0.729$ & $31.983$ & $\mathbf{0.251}$ & $\mathbf{9.150}$ \\
         \cmidrule(lr){1-6}
        \multirow{4}{*}{Conan}
        & Animate-Anyone & $0.768$ & $30.833$ & $0.311$ & $16.938$ \\
        \cmidrule(lr){2-6}
        & MYA & $0.815$ & $33.917$ & $0.231$ & $9.823$ \\
        \cmidrule(lr){2-6}
        & GAN Baseline & $0.842$ & $33.351$ & $0.202$ & $15.012$ \\
        \cmidrule(lr){2-6}
         & Ours & $0.821$ & $\mathbf{34.106}$ & $\mathbf{0.161}$ & $\mathbf{8.060}$ \\
        \bottomrule 
    \end{tabular}
    }

    % \vspace{5mm}
    \caption{
    Quantitative comparison on the in-domain test set.
    }
    \label{tab:table1}
    % \vspace{-5mm}
\end{table}

\begin{table}[t!]
    \caption{
    Quantitative comparison on the out-of-domain test set.
    }
    \centering
    \resizebox{0.95\linewidth}{!}{
    
    \begin{tabular}{l c c c c c}
        \toprule

        \textbf{Speaker} & \textbf{Method} & \textbf{SSIM} $(\uparrow)$ & \textbf{PSNR} $ (\uparrow)$ & \textbf{LPIPS} $(\downarrow)$ & \textbf{FVD} $(\downarrow)$
        
        \\
        \cmidrule(lr){1-6}
        \multirow{2}{*}{Oliver}
        & GAN Baseline & $0.715$ & $30.762$ & $0.315$ & $23.092$ \\
        \cmidrule(lr){2-6}
         & Ours & $\mathbf{0.744}$ & $\mathbf{31.895}$ & $\mathbf{0.254}$ & $\mathbf{12.819}$ \\
         \cmidrule(lr){1-6}
        \multirow{2}{*}{Chemistry}
        & GAN Baseline & $0.707$ & $30.862$ & $0.288$ & $24.507$ \\
        \cmidrule(lr){2-6}
         & Ours & $\mathbf{0.729}$ & $\mathbf{31.983}$ & $\mathbf{0.245}$ & $\mathbf{8.890}$ \\
         \cmidrule(lr){1-6}
        \multirow{2}{*}{Conan}
        & GAN Baseline & $0.821$ & $32.121$ & $0.216$ & $16.939$ \\
        \cmidrule(lr){2-6}
         & Ours & $\mathbf{0.843}$ & $\mathbf{33.209}$ & $\mathbf{0.183}$ & $\mathbf{9.611}$ \\
        \bottomrule 
    \end{tabular}
    }

    % \vspace{2mm}
    
    \label{tab:table2}
    % \vspace{-5mm}
\end{table}

% \myparagraph{Quantitative Comparison.}
\subsubsection{Quantitative Comparison.}
Table \ref{tab:table1} and Table \ref{tab:table2} show the comparison results of our method with the baseline GAN model and two diffusion model-based methods \cite{animateanyone, makeyouranchor} on two test sets in terms of image-level and video-level metrics. On in-domain data, our method outperforms the baseline in all metrics except for SSIM. The slight lag in the SSIM metric is because GANs-based models have stronger capabilities in restoring image brightness and contrast than diffusion models fine-tuned with a small amount of data. However, qualitative experiments show that our method significantly surpasses the baseline in the quality of hand generation. On out-of-domain data, our method comprehensively leads over the baseline. This demonstrates that the videos generated by our method surpass those produced by the baseline method in image quality. On both test sets, our method is generally able to outperform other diffusion model-based methods, and our method is more effective in model and training.
% attributable to the pretrained diffusion model's robust prior knowledge on body shifts and the various positions in which hands appear.

% \begin{figure*}[t!]
%     \centering
%     \includegraphics[width=0.8\linewidth]{images/qualitive1_2.pdf}
%     % \vspace{-10pt}
%     \caption{
%     Qualitative comparison between our method and the baseline. 
%     In the in-domain test set (upper half), our method outperforms the baseline by generating clearer hands and more realistic facial expressions. In the out-of-domain test set (lower half), the baseline model fails to adapt to body shifts and new hand positions relative to the training set, resulting in incorrect appearances, while our method consistently generates accurate and high-quality images.
%     }
%     \label{fig:4_qualitive_domain}
%     % \vspace{-10pt}
% \end{figure*}

\begin{figure}[t!]
    \centering
    \includegraphics[width=0.9\linewidth]{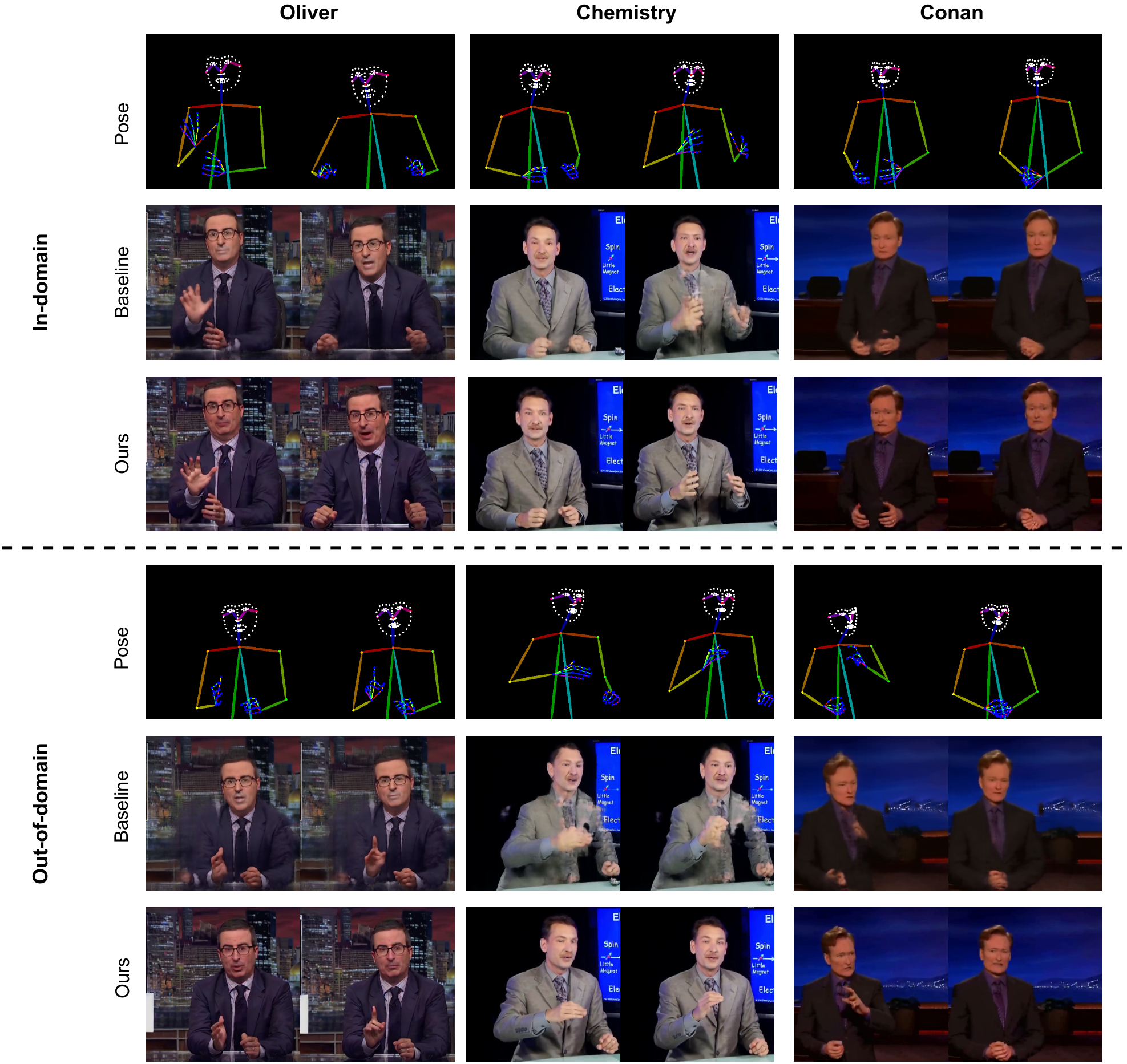}
    \vspace{-10pt}
    \caption{
    Qualitative comparison between our method and the GAN baseline. 
    In the in-domain test set (upper half), our method outperforms the baseline by generating clearer hands and more realistic facial expressions. In the out-of-domain test set (lower half), the baseline model fails to adapt to body shifts and new hand positions relative to the training set, resulting in incorrect appearances, while our method consistently generates accurate and high-quality images.
    }
    \label{fig:4_qualitive_domain}
    \vspace{-10pt}
\end{figure}

% \myparagraph{Qualitative Comparison.}
\subsubsection{Qualitative Comparison with GAN Basiline}
In Figure \ref{fig:4_qualitive_domain}, we compare the video quality generated by our method and the baseline method across all three subjects. Our method produced more natural and clearer facial expressions and more defined hands on both test sets. This advantage stems from the pre-trained SD and ControlNet's robust prior knowledge of body shifts and the corresponding shapes when hands appear in different positions. Notably, in the out-of-domain data, the results for Chemistry and Oliver (Figure \ref{fig:4_qualitive_domain}, 5th row, 1-4th columns) show that the baseline method generated blurry bodies or even missing parts. This issue arose because their body positions were fixed in the images of the training set, whereas there were shifts in the test set. This indicates that GANs only learned to generate images with fixed positions, while our method successfully generated correct appearances and hand poses after position changes. In section \ref{sec:camera}, we will further discuss the adaptability of the two methods to variations in subject magnification.

% \begin{figure}[htbp]
% \centering % 让图片居中显示
% \includegraphics[width=\linewidth]{images/zoom_in_1.pdf} 
% \caption{xxx}
% \label{fig:6_camera_adjust_1}
% \end{figure}
\begin{figure}[t!]
    \centering
    \includegraphics[width=0.9\linewidth]{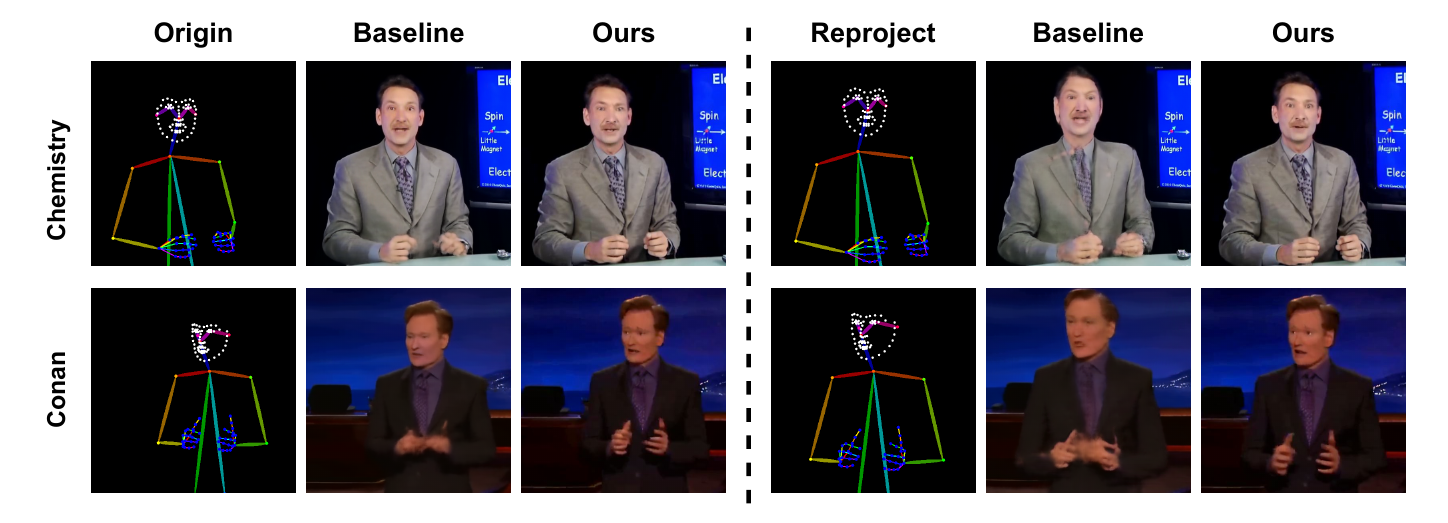}
    
        \vspace{-10pt}
    \caption{
    Generation results on the original skeleton map (left side) and on the skeleton map after increasing the camera focal length (right side) for both the Baseline method and our method. After the skeleton maps were enlarged, our method not only continued to generate clear hands but also produced correct head and body shapes compared to the Baseline.
        \vspace{-10pt}
    }
    \label{fig:6_camera_adjust_1}
\end{figure}

\subsection{Comparison of Camera Changes}
\label{sec:camera}
This section further discusses the adaptability of two methods when the camera parameters change, specifically when the input skeleton maps are enlarged relative to those in the training set, questioning whether both methods can adapt to this change and generate correct outcomes. Figure \ref{fig:6_camera_adjust_1} presents that when the camera's focal length increases, meaning the skeleton is enlarged, GANs cannot generate accurate images, manifesting incorrect body and head proportions. In contrast, our method succeeds in generating the correct appearance of the enlarged characters, proving its adaptability to changes in body position and size.

\begin{figure}[t!]
\centering %
\includegraphics[width=0.8\linewidth]{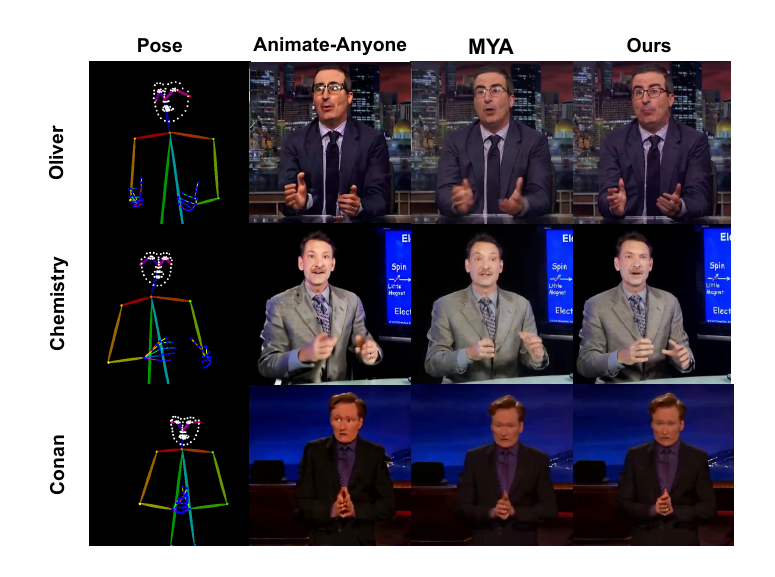} 

\vspace{-10pt}
\caption{
 We compared our method with the results generated by Animate-Anyone and MYA. The skeleton maps are derived from the Ground Truth of the out-of-domain test set. Our method produces clearer hands and more expressive facial features.
}
\label{fig:7_comp_disco}
% \vspace{-10pt}
\end{figure}

\begin{table}[t!]
    \caption{
    Ablation experiments on the main designs.
    }
    \centering
    \resizebox{0.95\linewidth}{!}{
    
    \begin{tabular}{l c c c c c}
        \toprule

        \textbf{Design} & \textbf{Method} & \textbf{SSIM} $(\uparrow)$ & \textbf{PSNR} $ (\uparrow)$ & \textbf{LPIPS} $(\downarrow)$ & \textbf{FVD} $(\downarrow)$
        
        \\
        \cmidrule(lr){1-6}
        \multirow{2}{*}{Skeleton map}
        & OpenPose & $0.703$ & $30.429$ & $0.295$ & $18.762$ \\
        \cmidrule(lr){2-6}
         & Refined (Ours) & $\mathbf{0.744}$ & $\mathbf{32.336}$ & $\mathbf{0.235}$ & $\mathbf{10.664}$ \\
         \cmidrule(lr){1-6}
        \multirow{2}{*}{Attention}
        & Self-Attention & $0.734$ & $32.158$ & $0.268$ & $10.737$ \\
        \cmidrule(lr){2-6}
         & Face Enhance Attnetion (Ours) & $\mathbf{0.744}$ & $\mathbf{32.336}$ & $\mathbf{0.235}$ & $\mathbf{10.664}$ \\
         \cmidrule(lr){1-6}
        \multirow{2}{*}{Temporal consistency}
        & Temporal module & $0.749$ & $32.175$ & $0.221$ & $11.064$ \\
        \cmidrule(lr){2-6}
         & Temporal Inference (Ours) & $0.744$ & $\mathbf{32.336}$ & $\mathbf{0.235}$ & $\mathbf{10.664}$ \\
        \bottomrule 
    \end{tabular}
    }

    \label{tab:table2}
    % \vspace{-5mm}
\end{table}

\subsection{Qualitative Comparison with Diffusion-based Method}
\label{sec:comp_diff}
In this section, we conduct a qualitative comparison of our method against Make-Your-Anchor \cite{makeyouranchor} and Animate-Anyone \cite{animateanyone}. We train with the original repository of \cite{makeyouranchor}, and for \cite{animateanyone} we train with the MooreThreads' implementation.

Figure \ref{fig:7_comp_disco} illustrates a visual quality comparison of generated images across three characters. Our method outperforms other methods in terms of hand quality and face clarity, which shows the effectiveness of refined 2D skeleton map and Face Enhance Attention.

\subsection{Ablation Study}

We conduct ablation studies to verify the effectiveness of the proposed methods. This section mainly evaluated three designs: 1) using skeleton maps extracted by OpenPose or refined 2D skeleton maps proposed by us. 2) Face Enhance Attention. 3) using temporal inference or training temporal module. The results are shown in Table 3, and we can see that our proposed method can achieve better performance while avoiding additional training. Note that the metrics in the table are averages on the Oliver and Chemistry test sets.

\section{Discussion}
\label{conclusion}

In this paper, we have introduced a diffusion-based system for generating co-speech gesture videos driven by audio inputs. Our method shows better generalization and clear hand poses than conditional GAN-based methods for video generation, and our model structure and training strategy are simpler and do not rely on large-scale pre-training compared to recent diffusion model-based methods. We have evaluated our method on the TALKSHOW dataset for a variety of audio inputs and character types. We plan to evaluate our method on larger datasets with more characters.

\bibliographystyle{IEEEbib}
\bibliography{aaai25}

\vspace{12pt}
\end{document}